\documentclass[runningheads]{llncs}
\usepackage{graphicx}
\usepackage{booktabs} 
\usepackage{tabularx}
\usepackage{arydshln} 
\usepackage{enumerate}
\usepackage{amssymb}
\usepackage{amsmath}
\usepackage{pgfplots}
\usepgfplotslibrary{dateplot}
\usepgfplotslibrary{groupplots}
\usepgfplotslibrary{fillbetween}
\usepgfplotslibrary{statistics}
\usepackage{mathtools}
\usepackage{nicefrac}
\usepackage[binary-units=true]{siunitx}
\usepackage{algorithmicx}

\usepackage{algorithm}
\usepackage{algpseudocode}

\makeatletter
\renewcommand*\env@matrix[1][\arraystretch]{%
  \edef\arraystretch{#1}%
  \hskip -\arraycolsep
  \let\@ifnextchar\new@ifnextchar
  \array{*\c@MaxMatrixCols c}}
\makeatother

\newcommand\defeq{\stackrel{\mathclap{\normalfont\mbox{\scriptsize def}}}{=}}

\begin{document}
\title{Learning Precise Spike Timings with\\Eligibility Traces}
\titlerunning{Learning Precise Spike Timings with Eligibility Traces}

%
%
\author{
	Manuel Traub\inst{1}\and
	Martin V. Butz\inst{1}\and
	{R.~Harald} Baayen\inst{2}\and
	Sebastian Otte\inst{1}
}
\authorrunning{M. Traub et al.}
%
\institute{
	University of T\"ubingen -- Neuro-Cognitive Modeling Group,\\ 
	Sand 14, 72076 T\"ubingen, Germany
	\and
	University of T\"ubingen -- Quantitative Linguistics,\\
	Wilhelmstr. 19, 72074 T\"ubingen, Germany
}
\maketitle              

\begin{abstract}
Recent research in the field of spiking neural networks (SNNs) has shown that recurrent variants of SNNs, namely \emph{long short-term SNNs} (LSNNs), can be trained via error gradients just as effective as LSTMs. 
The underlying learning method (e-prop) is based on a formalization of eligibility traces applied to \emph{leaky integrate and fire} (LIF) neurons. 
Here, we show that the proposed approach 
cannot fully unfold \emph{spike timing dependent plasticity} (STDP). 
As a consequence, this limits in principle the inherent advantage of SNNs, that is, the potential to develop codes that rely on precise relative spike timings.
We show that STDP-aware synaptic gradients naturally emerge within the eligibility equations of e-prop when derived for a slightly more complex spiking neuron model, here at the example of the Izhikevich model.
We also present a simple extension of the LIF model that provides similar gradients.
In a simple experiment we demonstrate that the STDP-aware LIF neurons can learn precise spike timings from an e-prop-based gradient signal.
\end{abstract}

\section{Introduction}
\label{section:introduction}

\emph{Spike Timing Dependent Plasticity} (STDP) is assumed to be a fundamental learning principle in the brain \cite{caporale2008spike}.
It is considered a prerequisite for developing temporal codes in which precise relative spike timings are key, thus going beyond plain rate coding.
Accordingly, STDP is based on temporal correlations between presynaptic and postsynaptic neural activities. 
%
%
Indeed, it has been shown that such a Hebbian-like form of synaptic plasticity is at play in the visual cortex of primates \cite{huang2014associative}. 

Recently, (non-supervised) STDP based learning rules were successfully applied to train deep (non-recurrent) convolutional SNNs for image recognition \cite{kheradpisheh2018stdp,mozafari2018first}.
In terms of effectiveness, though, supervised back-propagation-like approaches seem to be more promising \cite{kheradpisheh2019s4nn}, even though spiking neurons are not differentiable per se, such that an error gradient signal can only be approximated.

Recently, it has been shown that even \emph{back-propagation through time} (BPTT) \cite{werbos_backpropagation_1990} can be applied for training recurrent SNNs \cite{bellec2018long}.
Bellec et al. demonstrated that SNNs, specifically a variant that is called \emph{long short-term SNN} (LSNN), for the first time can reach the performance of the well-known LSTM \cite{hochreiter_long_1997}.
Moreover, the mathematical approach of learning in SNNs has led to the derivation of a biologically plausible learning rule called \emph{e-prop}, which approximates BPTT
\cite{bellec2019biologically} by means of a formalization of \emph{eligibility traces}.

Bellec et al. established a link between e-prop and (biological) synaptic plasticity \cite{bellec2019solution,bellec2019eligibility}.
It appears, however, that STDP can only fully arise within eligibility trace-based learning, when the neuron model provides a negated gradient signal in the case when a presynaptic spike arrives too late, i.e., shortly after a postsynaptic spike. This does not happen in the LIF model proposed in \cite{bellec2019eligibility}, cf. \autoref{section:stdplif} for further details.

The contributions of this paper are as follows. 
We show that  STDP emerges within e-prop when it is derived using a more complex neuron model, which adequately incorporates a refractory period.
This is exemplarily shown for the Izhikevich model \cite{izhikevich2003simple}. 
Moreover, we present an adjustment of the basic LIF model used in LSNNs in order to produce the same STDP behavior within the e-prop framework.


\section{Background}
Experimental data suggests that the brain solves the \emph{temporal credit assignment problem} by combining local eligibility traces, which maintain information about individual synapses' activation histories, with neuromodulator-based reward signals \cite{gerstner2018eligibility}. 
The e-prop algorithm \cite{bellec2019biologically} adapts this principle by factorizing the error gradients from BPTT
into a sum of products between local eligibility traces and online learning signals.
\begin{align}\label{eq:eprop}
\frac{d E}{d w_{i,j}} &= \sum_t L_j^t e_{i,j}^t \nonumber\\
&=\sum_{t} \frac{d E}{d z_j^t}
e_{i,j}^t
\end{align}
Here, the learning signal ($L_j^t$) represents the global error information of the postsynaptic spike,
whereas the eligibility trace ($e_{i,j}^t$) captures the local information available at the synapse. $z_{j}^{t}$ refers to the (spiking) output of neuron $j$ at time step $t$.

The eligibility trace is furthermore a product of pre- and postsynaptic information, but does not include any error gradient information:
\begin{equation}\label{eq:eproptrace}
e_{i,j}^t = \frac{\partial z_j^{t}}{\partial \textbf{s}_j^{t}} \boldsymbol \epsilon_{i,j}^t
\end{equation} 

Specifically, it is the product of a \emph{pseudo derivative}, replacing the non-existing derivative of the spiking function $\nicefrac{\partial z_j^{t}}{\partial \textbf{s}_j^{t}}$ and the presynaptic activity flow accumulated within an eligibility vector $\boldsymbol \epsilon_{i,j}^t$. The latter is computed forward through time:
\begin{equation}\label{eq:eligibility_vector}
\boldsymbol \epsilon_{i,j}^t = \boldsymbol \epsilon_{i,j}^{t-1} \frac{\partial \textbf{s}_j^{t}}{\partial \textbf{s}_j^{t-1}} + \frac{\partial \textbf{s}_j^{t}}{\partial w_{i,j}}
\end{equation}
where $\mathbf{s}_{j}^{t}$ refers to the current state of a neuron containing, for instance (depending on the model), its action potential and possibly further adaptive parameters.

\section{STPD with Izhikevich Neurons}
The Izhikevich neuron \cite{izhikevich2003simple} is a precise, but 
computationally cheap model of a biological neuron that uses two parameterizable differential 
equations. It is particularly more complex than the simple LIF model, but, more importantly, explicitly models the refractory period of the neuron.

\subsection{Izhikevich Model and Eligibility Trace}
The dynamics of the Izhikevich neuron are described with the following differential equations. 

\begin{equation}\label{eq:izhikevich1}
v' = 0.04 v^2 + 5 v + 140 -u + I
\end{equation} 
\begin{equation}\label{eq:izhikevich2}
u' = 0.004 v - 0.02 u
\end{equation} 
Here $v$ is the membrane voltage and $u$ is a recovery variable, which controls the refractory period
of the Izhikevich neuron. $I$ is the current input to the neuron.

Once the membrane voltage crosses 30mV, 
a spike is emitted and $v$ and $u$ are reset as described in \autoref{alg:example}.
\begin{algorithm}[b]
   \caption{Izhikevich neuron reset}
   \label{alg:example}
\begin{algorithmic}
\If{$v < 30mv$}
    \State $v \gets -65mV$
    \State $u \gets u + 2$
\EndIf
\end{algorithmic}
\end{algorithm}

In order to derive an eligibility trace for Izhikevich neurons, the equations 
\eqref{eq:izhikevich1} and \eqref{eq:izhikevich2} have to be modeled in discrete time steps,  and also the reset has to be modeled within the new equations. To accomplish this built-in reset, the variables $\tilde{v}_j^t$ and $\tilde{u}_j^t $ are introduced, replacing $v$ and $u$ in the standard equations.
\begin{equation}\label{eq:reset_v}
\tilde{v}_j^t = v_j^t - (v_j^t + 65) z_j^t
\end{equation}
\begin{equation}\label{eq:reset_v}
\tilde{u}_j^t = u_j^t + 2 z_j^t
\end{equation}

Here the binary variable $z_j^t$, resets $\tilde{v}_j^t$ and $\tilde{u}_j^t$ after a spike of neuron $j$ at time step $t$.
Now $v$ and $u$ can be computed in discrete time steps using Euler integration with a 
constant step size of $\delta t$.
\begin{equation}
 v_j^{t+1} = \tilde{v}_j^t + \delta t (0.04 (\tilde{v}_j^t)^2 + 5\tilde{v}_j^t + 
             140 - \tilde{u}_j^t + I_j^t) 
\end{equation}  
\begin{equation}
 u_j^{t + 1} = \tilde{u}_j^t + \delta t (0.004 \tilde{v}_j^t - 0.02 \tilde{u}_j^t)
\end{equation}  

The hidden state of Izhikevich neurons is then 
defined as a two-dimensional vector containing $v_j^t$ and $u_j^t$.
\begin{equation}
\textbf{s}_j^t = 
\begin{pmatrix}[1.4] 
v_j^t \\
u_j^t 
\end{pmatrix}
\end{equation}  

To finally derive the eligibility vector \eqref{eq:eligibility_vector},
the derivative of the next hidden state $s_j^{t+1}$
by the current state $s_j^t$ has to be computed. 
This state derivative can be expressed in the following form of a 2x2 matrix. 
\begin{equation}
\frac{\partial \textbf{s}_j^{t+1}}{\partial \textbf{s}_j^t} 
= 
\begin{pmatrix}[2] 
	\frac{\partial v_j^{t+1}}{\partial v_j^t} & \frac{\partial v_j^{t+1}}{\partial u_j^t} \\ 
	\frac{\partial u_j^{t+1}}{\partial v_j^t} & \frac{\partial u_j^{t+1}}{\partial u_j^t} 
\end{pmatrix}\\
\end{equation}  

The partial derivatives can be further simplified by taking into account that $z_j^t$ is a binary variable.

\begin{equation}
\begin{split}
	\frac{\partial v_j^{t+1}}{\partial v_j^t} &=
	1 - z_j^t\\
    &\phantom{=} + 0.08 \delta t (v_j^t - (v_j^t + 65) z_j^t)(1 - z_j^t)\\
    &\phantom{=} + 5 \delta t (1 - z_j^t)\\\\
    &=	1 - z_j^t + 0.08 \delta t v_j^t (1 - z_j^t)\\
    &\phantom{=} + 5 \delta t (1 - z_j^t) \\\\
    &= (1 - z_j^t)(1 + (0.08 v_j^t + 5) \delta t )
\end{split}
\end{equation}
\begin{equation}
	  \frac{\partial v_j^{t+1}}{\partial u_j^t} = -\delta t
\end{equation}  
\begin{equation}
	\frac{\partial u_j^{t+1}}{\partial v_j^t} = 0.004 \delta t (1 - z_j^t)
\end{equation}  
\begin{equation}
    \frac{\partial u_j^{t+1}}{\partial u_j^t} = 1 - 0.02 \delta t
\end{equation}  

Given this state derivative, the eligibility vector is computed in the following 
two-dimensional vector, which contains a voltage eligibility value in the first row and 
a refractory eligibility value in the second row:
\begin{equation}\label{eq:izhikevich_elv}
\boldsymbol \epsilon_{i,j}^{t+1} = \frac{\partial \textbf{s}_j^{t+1}}{\partial \textbf{s}_j^t} \cdot \boldsymbol \epsilon_{i,j}^{t} + 
\frac{\partial \textbf{s}_j^{t+1}}{\partial \theta_{ji}^{rec}} 
\end{equation}  
where
\begin{equation}
\begin{split}
	\epsilon_{i,j,v}^{t+1} =& (1 - z_j^t)(1 + (0.08 v_j^t + 5)\delta t )\epsilon_{i,j,v}^t \\
    & - \delta t \epsilon_{i,j,u}^t + \delta t z_i^t 
\end{split}
\end{equation}  
\begin{equation}
	\epsilon_{i,j,u}^{t+1} = 0.004 \delta t (1 - z_j^t) \epsilon_{i,j,v}^t + (1 - 0.02 \delta t ) \epsilon_{i,j,u}^t 
\end{equation}  

Here the recovery eligibility value $\epsilon_{i,j,u}^t$ approximates an exponential filter 
of the voltage eligibility.

The voltage eligibility itself is a function of the neuron's action potential that also accumulates 
presynaptic spikes and gets dampened by its exponential average (by the recovery eligibility).

Another insight from the equations is that, whenever the postsynaptic neuron spikes, 
the voltage eligibility vector is reset to the negative
recovery eligibility vector.

In order to derive the final eligibility trace, a pseudo-derivative for the Izhikevich neuron
is defined as follows:
\begin{equation}
h_j^t := \gamma ~ \exp(\frac{\min(v, 30) -30}{30})
\end{equation}  
where $\gamma$ is a damping factor. 

With this pseudo-derivative, the neuron spike $z_j^t$ derived by the hidden state $\textbf{s}_j^t$ is then
defined as:
\begin{equation}
\begin{pmatrix}[2]
\frac{z_j^t}{v_j^t} \\
\frac{z_j^t}{u_j^t}
\end{pmatrix} \defeq
\begin{pmatrix}[1.4]
 h_j^t \\
 0
\end{pmatrix}
\end{equation}  

The eligibility trace then simplifies to the pseudo-derivative times the voltage 
eligibility vector.
\begin{equation}
\begin{split}
e_{i,j}^{t+1} &= \frac{\partial z_j^{t+1}}{\partial \textbf{s}_j^{t+1}} \cdot \boldsymbol \epsilon_{i,j}^{t+1}\\
&=
\begin{pmatrix}[1.4] 
	h_j^{t+1} &
	0
\end{pmatrix} 
\begin{pmatrix}[1.4]
    \epsilon_{i,j,v}^{t+1}\\
    \epsilon_{i,j,a}^{t+1} 
\end{pmatrix} \\
&=
h_j^{t+1} \epsilon_{i,j,v}^{t+1}
\end{split}
\end{equation}

\subsection{Experimental Results}
In the following evaluation, our goal was to inspect the evolution of the derived eligibility traces.
Two Izhikevich neurons were weakly connected by a synapse and received random inputs to simulate behavior within a greater network. To specifically investigate the influence of the eligibility trace on the gradient, a constant positive learning signal was used.
The gradient was calculated based on \eqref{eq:eprop}. 

\begin{figure}[t!]
    \centering
    \includegraphics[width=\linewidth]{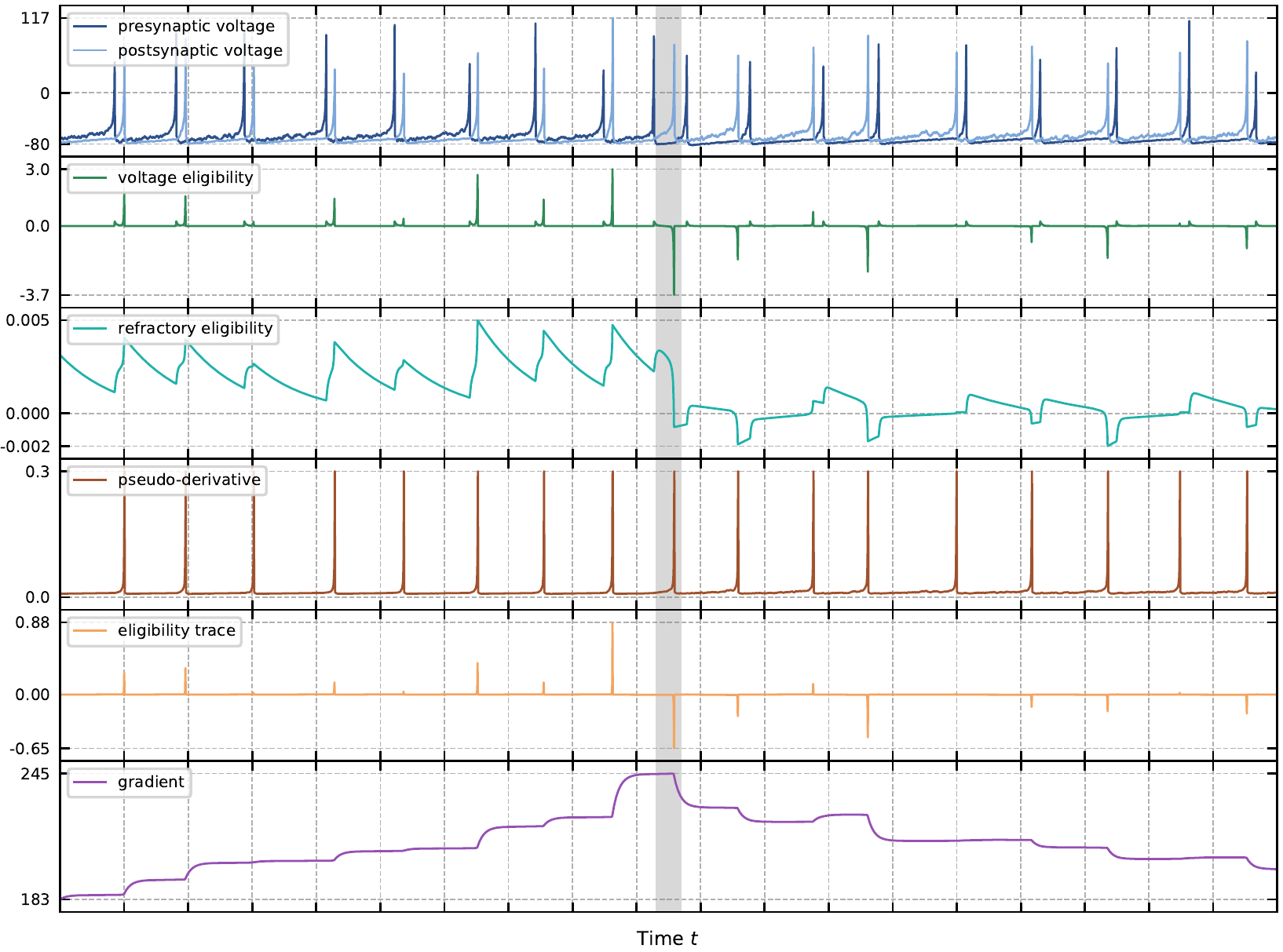}
    \caption{Simulation of two connected Izhikevich neurons that exhibit 
        a positively rewarded STDP behavior followed by a negatively rewarded one.
        As a result, a gradient 
        computed with a constant positive learning signal increases during positively rewarded STDP
        and decreases during negatively rewarded STDP.}
    \label{fig:izhikevich}
\end{figure}

In the simulation shown in \autoref{fig:izhikevich}, first an artificial strengthening STDP behavior is introduced by using an overall lower random input current to the output neuron.
To ensure that an output spike (postsynaptic neuron) fires shortly after an input spike (presynaptic neuron) we steadily increased the random input current for the output neuron after the input neuron spiked.
In the second part of the simulation this behavior is reversed (presynaptic spike before postsynaptic spike) in order to produce a weakening STDP behavior, as shown in Algorithm \ref{alg:izhikevich_eprop}.
Here $U(\alpha, \beta)$ represents a random variable drawn uniform from the interval $(\alpha, \beta)$ and $t_{z_i}, t_{z_o}$ are the spike times for the last input or output spike.

\begin{algorithm}[htb]
	\caption{Izhikevich e-prop simulation}
	\label{alg:izhikevich_eprop}
	\begin{algorithmic}
		\If{$t < T \cdot 0.45 $}
		\State $I_i^t \gets U(1, 15)$
		\State $I_o^t \gets U(1, 5)$
		\If{$t_{z_i} > t_{z_o}$}
		\State $I_o^t \gets U(0, 1) (t - t_{z_i})$
		\EndIf
		\Else
		\State $I_i^t \gets U(1, 5)$
		\State $I_o^t \gets U(1, 15)$
		\If{$t_{z_o} > t_{z_i}$}
		\State $I_i^t \gets U(0, 1) (t - t_{z_o})$
		\EndIf
		\EndIf
	\end{algorithmic}
\end{algorithm}

As can be seen in the diagram, the desired STDP behavior of the synapse is clearly reflected within the gradient, 
which for the first half of the simulation increases, and then decreases.
The eligibility trace and the eligibility vector also reflect the introduced STDP behavior.
They have only positive values during the STDP strengthening phase, but they also take on negative
values during the STDP weakening phase.

\section{STDP with LIF Neurons}\label{section:stdplif}
While our experimental results confirm the desired general tendency to reflect full STDP behavior in gradients based on the eligibility traces of Izhikevich neurons, plain LIF neurons do not show this behavior.
We now detail the reason for this lack and introduce negative eligibilities to induce effective connection weakening.

\subsection{LIF Model}
Considering the definition of the dynamics of the LIF neuron
\begin{equation}\label{eq:lif_original}
v_j^{t + 1} = \alpha v_j^t + I_j^t - z_j^t v_{thr}
\end{equation}
Equation~\eqref{eq:lif_original} shows the action potential dynamics of a LIF neuron in discrete timesteps, as defined in \cite{bellec2019biologically}. The neuron integrates over the weighted sum of incomming
spikes $I_j^t$; $\alpha < 1 $ controls the leakage. Once a LIF neuron spikes, its action potential is reset by subtracting the value of the spike threshold. 

A neuron spike is modeled by the Heaviside step function.
The neuron is prohibited from spiking during a fixed refractory period after the last spike.
\begin{equation}\label{eq:lif_spike}
z_j^t = 
    \begin{cases}
        0, & \text{if} ~ t - t_{z_j} < \delta t_{ref} \\
        H(v_j^t - v_{thr}), & \text{otherwise}
    \end{cases}
\end{equation}
Here \eqref{eq:lif_spike} $t_{z_j}$ represents the most recent spike time of neuron $j$
and $\delta t_{ref}$ denotes the length of the refractory period.

Seeing that also the \emph{pseudo derivative} \eqref{eq:lif_pseudoderivative} is set to zero during this refractory period:
\begin{equation}\label{eq:lif_pseudoderivative}
h_j^t := 
    \begin{cases}
        0, & \text{if} ~ t - t_{z_j} < \delta t_{ref} \\
        \gamma ~ \text{max}(0, 1 - |\frac{v_j^t - v_{thr}}{v_{thr}}|), & \text{otherwise}
    \end{cases},
\end{equation}
no weakening STDP based gradients can unfold in the e-prop equations using standard LIF neurons.

\subsection{STDP-LIF Model and Eligibility Trace}
Whereas an STDP influenced gradient cannot be observed with the standard equations
for LIF or adaptive LIF (ALIF) neurons within an LSNN \cite{bellec2018long}, by slightly modifying the original LIF
formulation from \eqref{eq:lif_original}, a clear STDP based eligibility trace emerges.

In order to compute such an eligibility trace reflecting the STDP behavior of the synapse connecting two LIF neurons, the LIF equation has to be slightly altered into, what we from now on will refer to, an STDP-LIF neuron:
\begin{equation}\label{eq:alternative_lif}
 v_j^{t+1}   = \alpha v_j^t + I_j^t - z_j^t \alpha v_j^t - z_j^{t - \delta t_{\text{ref}}} \alpha v_j^t 
\end{equation}  
Instead of using a soft reset at a fixed threshold, in 
\eqref{eq:alternative_lif} the STDP-LIF neuron is hard reset to zero whenever it spikes,
and whenever its refractory period ends.
Since the reset now does include the voltage $v_j^t$ as a factor,
the spike is now included in the hidden state derivative
\begin{equation}
\begin{split}
	\frac{\partial v_j^{t+1}}{\partial v_j^t} &= \alpha - z_j^t \alpha - \alpha z_j^{t - \delta t_{ref}} \\
    &= \alpha (1 - z_j^t - z_j^{t - \delta t_{ref}})
\end{split}
\end{equation}  
and hence also in the computation of the eligibility trace
\begin{equation}
\begin{split}
\boldsymbol \epsilon_{i,j}^{t+1} &= \frac{\partial \textbf{s}_j^{t+1}}{\partial \textbf{s}_j^t} \cdot \boldsymbol \epsilon_{i,j}^{t} + 
\frac{\partial \textbf{s}_j^{t+1}}{\partial \theta_{ji}^{rec}} \\
&=  \alpha (1 - z_j^t - z_j^{t - \delta t_{ref}}) \boldsymbol \epsilon_{i,j}^{t} + z_i^t
\end{split}
\end{equation}  
\begin{equation}
\begin{split}
e_{i,j}^{t+1} &= \frac{\partial z_j^{t+1}}{\partial \textbf{s}_j^{t+1}} \cdot \boldsymbol \epsilon_{i,j}^{t+1} \\
&= 	h_j^{t+1} \boldsymbol \epsilon_{i,j}^{t+1}
\end{split}
\end{equation}  

As a result, the eligibility trace is reset after a spike and after the refractory
period. 

This reset behavior allows us now to create an eligibility trace that reflects the STDP behavior of its synapse
by using a constant negative pseudo-derivative during the refractory period 
and otherwise leave the neuron dynamics unchanged. 
\begin{equation}\label{eq:stdp_lif_pseudoderivative}
h_j^t := 
    \begin{cases}
        -\gamma, & \text{if} ~ t - t_{z_j} < \delta t_{ref} \\
        \gamma ~ \text{max}(0, 1 - |\frac{v_j^t - v_{thr}}{v_{thr}}|), & \text{otherwise}
    \end{cases}
\end{equation}
As a result, any incoming spike during the refractory period produces a negative 
eligibility trace that persists for the time of the refractory period and 
has a negative influence on the gradient. 

\begin{figure}[t!]
    \centering
    \includegraphics[width=\textwidth]{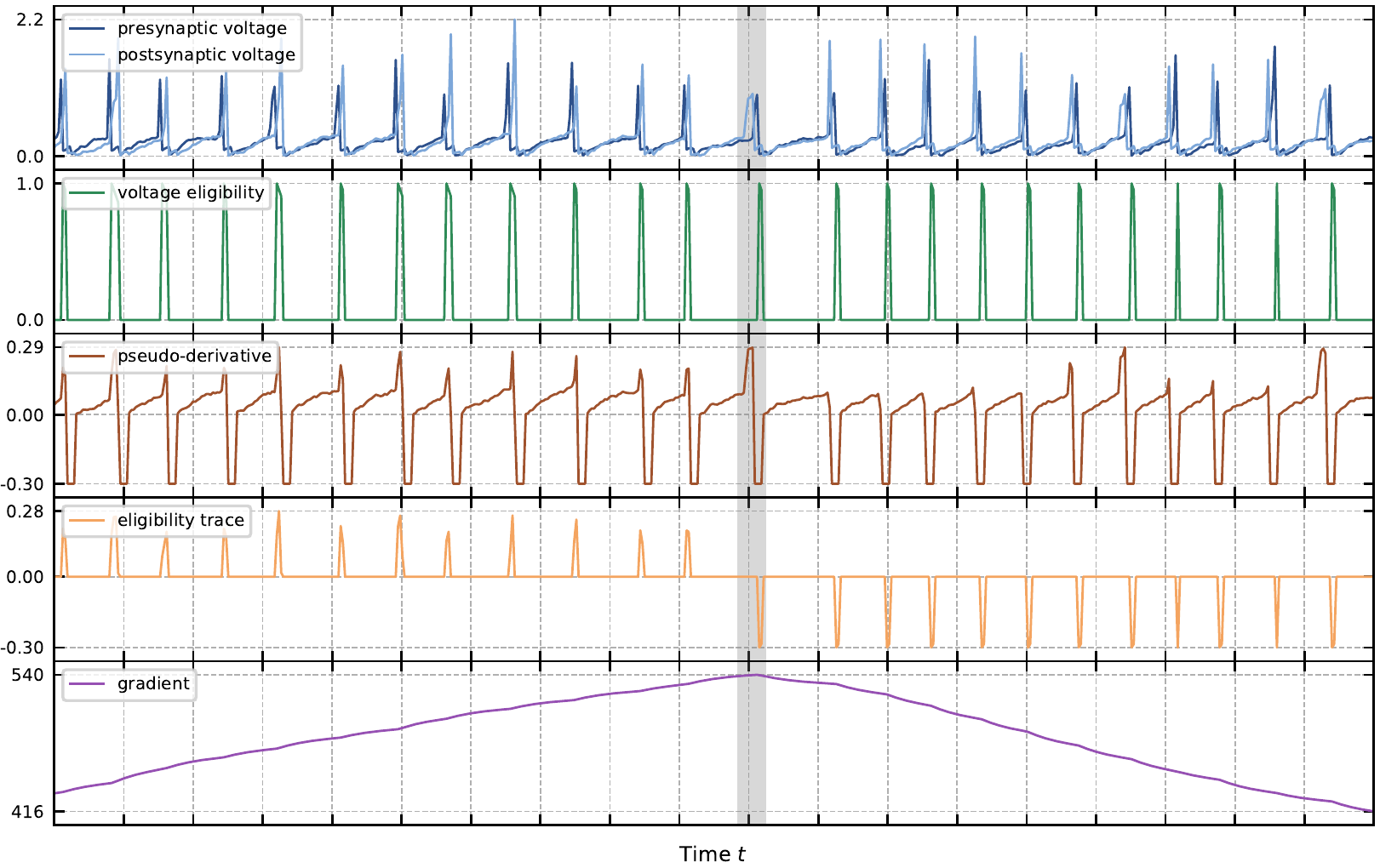}
    \vspace{-0.5cm}
    \caption{Simulation of two connected LIF neurons that exibit 
        a positively rewarded STDP behavior followed by a negatively rewarded one.
        As a result, a gradient 
        computed with a constant positive learning signal increases during positively rewarded STDP
        and decreases during negatively rewarded STDP.}
    \label{fig:lif}
\end{figure}

\subsection{Experimental Results}
The STDP influenced gradient is shown in \autoref{fig:lif}, in which two connected STDP-LIF neurons are simulated similarly to the Izhikevich neuron simulation in Algorithm~\ref{alg:izhikevich_eprop}.
Strengthening and weakening STDP events can be clearly identified  
in the eligibility trace, directly influencing the resulting gradient.
Thus, by the above simple modifications of the LIF model, one can derive eligibility traces for STDP-LIF neurons
that reflect the STDP behavior of the underlying synapse. 

Since the gradient calculation 
using eligibility traces in combination with a back-propagated learning signal \eqref{eq:eprop} is mathematically equivalent to normal BPTT, it follows that BPTT itself facilitates STDP behavior.

\subsection{Learning Precise Timing}
In a simple additional experiment we evaluated the performance of STDP-LIF neurons in comparison to standard LIF neurons
by learning an LSNN with 16 hidden neurons, one input neuron, and one output neuron to approximate a function based on the timing of the input spike.
In this experiment the LSNN receives Poisson distributed spikes from the input neuron with an average spike rate of $25\,\text{Hz}$. The supervised target signal for the leaky readout neuron is then
\begin{equation}\label{eq:spike_input}
v_{target}^t = \frac{1}{1 + t - t_{in}}
\end{equation}

In this equation, $t_{in}$ is the spike time of the most recent input spike.
A good approximation of \eqref{eq:spike_input} can be learned when the input spikes are just forwarded to
the readout neuron, which can in turn approximate the shape of the function via its readout decay.

In the simulation, the spike threshold $v_{thr}$ for each neuron was set to  $0.5$ and the LSNN was trained over $1\,000$ epochs using a batch size of 16 and an initial learning rate of $0.003$, which is decayed every 100 epochs by multiplying it with $0.7$. The network was trained with Adam \cite{kingma_adam_2015} and no further regularization was used.

\begin{figure}[t!]
    \input{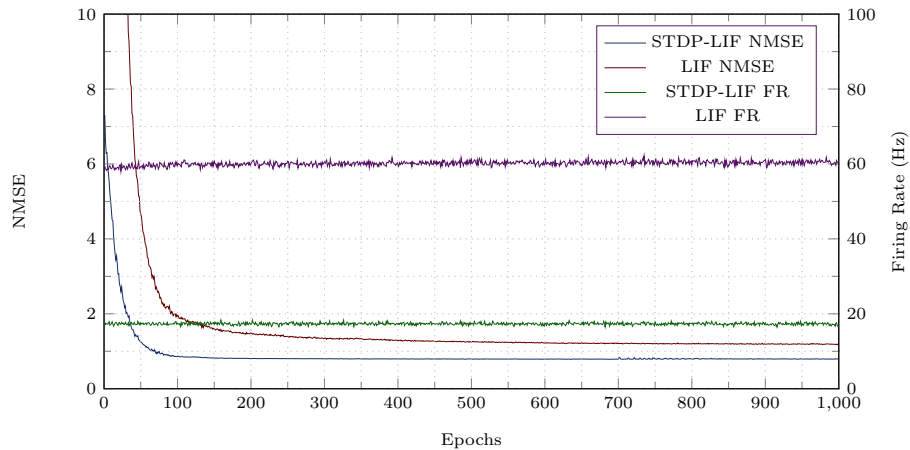}
    \vspace{-0.5cm}
      \caption{Spike timing test: STDP-LIF neurons manage to learn a simple function based on the inputs spike timing
               efficiently, while standard LIF neurons produce a lager error by approximating the function using a high firing rate.
               Graphs are averaged over 100 independent runs each.}
      \label{fig:spiketimingtest}
    \end{figure}
    
As can be seen in \autoref{fig:spiketimingtest},
the STDP-LIF neurons quickly manage to suppress the hidden connections and only forward the input spikes
to the readout neuron, while the LIF neurons on the other hand produce a lager error by approximating the function 
with a high firing rate.

\section{Conclusion}
In this paper we showed that STDP emerges within eligibility trace-based gradient signals in SNNs, given that the neuron model is sufficiently detailed. 
Specifically, it is crucial that the refractory period of postsynaptic neurons is taken into account. While this is not the case for regular LIF neurons, we demonstrated that by including eligibility traces derived from the well-known Izhikevich model, the gradient signal induces STDP behavior. The standard LIF model can also be refined such that it provides STDP. Equipped with STDP-aware gradient signals, we showed that learning precise spike timings becomes possible.

Seeing the recent advances in applying (recurrent) SNNs successfully, we will evaluate these mathematical insights very soon to harder benchmark problems 
like handwriting or speech recognition (TIMIT) and expect to achieve similar performances. The observed effect of reduced firing rates when using STDP-aware gradients could be particularly interesting for neuromorphic hardware implementations, in which sparse spiking behavior effectively benefits energy consumption.


%
\bibliographystyle{splncs04}
\bibliography{2020-STDPSNN}
\end{document}